\documentclass[letter, 10 pt, conference]{ieeeconf}  
\IEEEoverridecommandlockouts                              
\usepackage{graphics}    
\usepackage{times}       
\usepackage{amsmath}     
\usepackage{amssymb}     
\usepackage{graphicx}
\usepackage{subcaption}
\usepackage{hyperref}
\usepackage{xspace}
\usepackage{color}

\usepackage[font=small]{caption}

\def\secref#1{Sec.~\ref{#1}}
\def\figref#1{Fig.~\ref{#1}}
\def\tabref#1{Tab.~\ref{#1}}
\def\eqref#1{Eq.~(\ref{#1})}

\newcommand{\lidar}{LIDAR\xspace}
\newcommand{\channel}{\mathrm{c}}
\newcommand{\Channels}{\mathcal{C}}
\newcommand{\intensity}{\mathrm{intensity}}
\newcommand{\depth}{\mathrm{depth}}
\newcommand{\normal}{\mathrm{normal}}
\newcommand{\pinhole}{\mathrm{pinhole}}
\newcommand{\spherical}{\mathrm{spherical}}

\newcommand{\I}{\boldsymbol{\mathfrak{I}}}
\newcommand{\IPred}{\hat{\boldsymbol{\mathfrak{I}}} (\cM, \bX)}

\newcommand{\IChannel}[1][]{\boldsymbol{\mathfrak{I}}^\channel_{#1}}
\newcommand{\IChannelPred}[1][]{\hat{\boldsymbol{\mathfrak{I}}}^\channel_{#1} (\cM, \bX)}
\newcommand{\UV}{{(u~v)}^\mathrm{T}}
\newcommand{\Trans}{\top}
\newcommand{\ErrorPixel}{\be^{\channel}_{u,v} (\cM, \bX)}
\newcommand{\ErrorPoint}[1][\bX]{\be^{\channel}_{\bp} (#1)}
\newcommand{\ErrorPointConstant}{\breve \be^{\channel}_{\bp}}

\newcommand{\ModelVis}{\cM_{\mathrm{vis}}}

\newcommand{\ProjFuncName}{\proj}
\newcommand{\ProjFunc}[1][\bX\,\bp]{\proj(#1)}
\newcommand{\ProjFuncChannel}[1][]{\proj^{#1} (\bp)}

\newcommand{\MapFuncName}[1][\channel]{\map^{#1}}
\newcommand{\MapFunc}[1][\channel]{\map^{#1} (\bX,\bp)}
\newcommand{\MapFuncDer}[1][\channel]{\map^{#1} (\bX\oplus\bx,\bp)}

\newcommand{\Jp}{\bJ^{\channel}_\bp(\bX)}
\newcommand{\JpConstant}{\breve \bJ^{\channel}_\bp}
\newcommand{\JpConstantTrans}{\breve \bJ^{\channel \Trans}_\bp}
\newcommand{\Jmap}{\bJ^{\channel}_\map}
\newcommand{\Jimg}{\bJ^{\channel}_\img}
\newcommand{\Jproj}{\bJ_\proj}
\newcommand{\Jtf}{\bJ_\mathrm{tf}}

\newcommand{\RGBD}{RGBD }

\usepackage{amsopn}

\newcommand{\bv}{\mathbf{v}}

\newcommand{\bt}{\mathbf{t}}

\newcommand{\bD}{\mathbf{D}}
\newcommand{\bK}{\mathbf{K}}

\newcommand{\bH}{\mathbf{H}}
\newcommand{\bI}{\mathbf{I}}
\newcommand{\bP}{\mathbf{P}}
\newcommand{\bX}{\mathbf{X}}

\newcommand{\bR}{\mathbf{R}}

\newcommand{\brevep}{\breve{\mathbf{p}}}

\newcommand{\bJ}{\mathbf{J}}
\newcommand{\bZero}{\mathbf{0}}

\newcommand{\cM}{\mathcal{M}}

\newcommand{\ba}{\mathbf{a}}
\newcommand{\bb}{\mathbf{b}}

\newcommand{\be}{\mathbf{e}}

\newcommand{\bx}{\mathbf{x}}

\newcommand{\bn}{\mathbf{n}}

\newcommand{\bp}{\mathbf{p}}
\newcommand{\bDelta}{\mathbf{\Delta}}

\newcommand{\bDeltaR}{\mathbf{\Delta R}}
\newcommand{\bDeltax}{\mathbf{\Delta x}}
\newcommand{\bDeltat}{\mathbf{\Delta t}}

\newcommand{\proj}{\mathrm{proj}}
\newcommand{\map}{\mathrm{map}}
\newcommand{\range}{\mathrm{range}}

\newcommand{\img}{\mathrm{img}}
\newcommand{\norm}{\mathrm{norm}}
\newcommand{\vectorToTransform}{\mathrm{v2t}}

\newcommand{\bOmega}{\mathbf{\Omega}}

\DeclareMathOperator*{\argmin}{argmin}
\DeclareMathOperator*{\atantwo}{atan2}

\title{\LARGE \bf A General Framework for Flexible \\ Multi-Cue Photometric Point Cloud Registration}

\author{Bartolomeo Della Corte$^*$ \and Igor Bogoslavskyi$^*$ \and
  Cyrill Stachniss \and Giorgio Grisetti
  \thanks{$^*$~These two authors contribute equally to the work.}%
  \thanks{Igor and Cyrill are with the University of Bonn, Germany. Bartolomeo and
    Giorgio are with Sapienza University of Rome, Department of
    Computer, Control, and Management Engineering Antonio Ruberti,
    Rome, Italy.
  }%
  \thanks{This work has partly been supported by the EC under the
  grant number H2020-ICT-644227-Flourish and the DFG under the grant
  number FOR~1505: Mapping on Demand.
  }%
}

\begin{document}
\maketitle
\thispagestyle{empty}
\pagestyle{empty}

\begin{abstract}
  %
  The ability to build maps is a key functionality for the majority of
  mobile robots. A central ingredient to most mapping systems is the
  registration or alignment of the recorded sensor data.
  In this paper, we present a general methodology for photometric
  registration that can deal with multiple different cues.
  We provide examples for registering \RGBD as well as 3D
  \lidar data.
  In contrast to popular point cloud registration approaches such as
  ICP our method does not rely on explicit data association and
  exploits multiple modalities such as raw range and image data
  streams.  Color, depth, and normal information are handled in an
  uniform manner and the registration is obtained by minimizing the
  pixel-wise difference between two multi-channel images.
  We developed a flexible and general framework and implemented our
  approach inside that framework.  We
  also released our implementation  as  open source C++ code.
  The experiments show that our approach allows for an
  accurate registration of the sensor data
  without requiring an explicit data association or model-specific adaptations
  to datasets or sensors. Our approach  exploits
  the different cues in a natural and consistent way and the registration can
  be done at framerate for a typical range or imaging sensor.
\end{abstract}

\section{Introduction}
\label{sec:intro}

Most mobile robots need to estimate a map of their surroundings in
order to navigate. Thus, the task of registering the incoming sensor
data such as images or point clouds is an important building block for
most autonomous systems. This functionality is also of key importance
for estimating the relative motion of a robot through incremental
matching, often called visual odometry or laser-based odometry,
depending on the used sensing modality.

We investigate the problem of registering data from typical robotic
sensors such as the Kinect camera, a 3D \lidar such as a Velodyne laser
scanner, or similar in a general way without requiring special,
sensor-specific adaptations.  More concretely our goal is to provide a
general methodology to find the transformation that maximizes the
overlap between two measurements taken from the same scene.

To this extent, ICP is a popular strategy for registering point
clouds.  It proceeds by iteratively alternating two steps: data
association and transform estimation.  Data association computes pairs
of corresponding points in the two clouds, while transform estimation
calculates an isometry that applied to one of the two clouds minimizes
the distance between corresponding points. The  weakness of ICP lies
in the correspondence search as this step usually relies on  heuristics
and may introduce biases or gross errors.

\begin{figure}[t]
  \centering
  \includegraphics[width=.95\columnwidth]{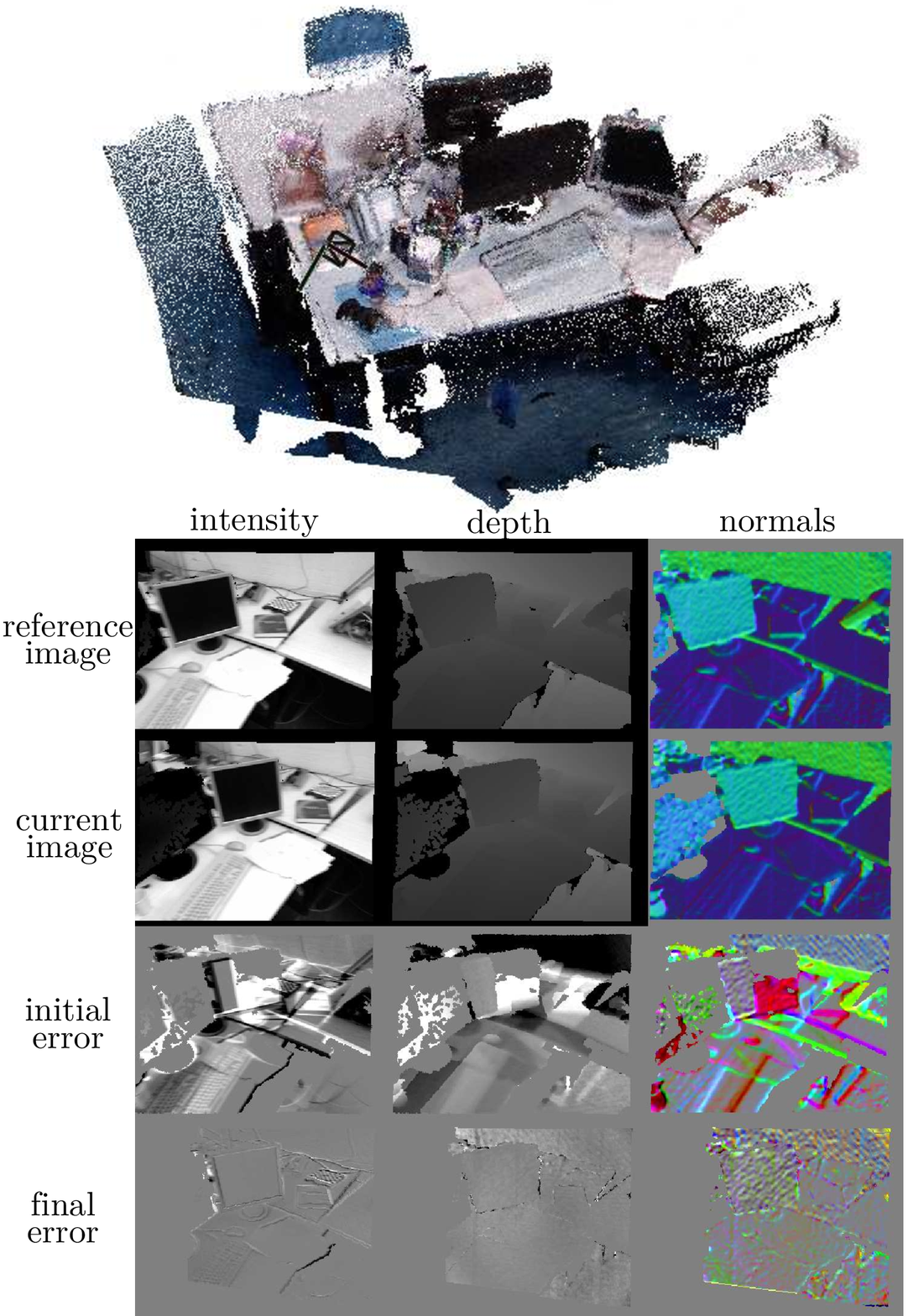}
  \caption{Our approach realized an effective multi-cue registration without requiring features nor an explicit data association between features or 3D points.}
  \label{fig:motivation}
\end{figure}

Over time, effective variants of ICP that exploit the structure of the
scene have been proposed~\cite{segal2009rss, serafin2015iros}.  Using
the structure either by relying on a point-to-plane or a plane to
plane metric has been shown to improve the performance of the
algorithm, especially in indoor/structured environments.
Kerl~\emph{et al.}~\cite{kerl2013icra} recently proposed Dense Visual
Odometry (DVO), an approach to register RGBD images by minimizing the
photometric distance. The idea is to find the location of a camera
within a scene such that the image captured at that location is as
close as possible to the measured RGBD image.  By exploiting the image
gradients, DVO does not require explicit data association, and it is able
to achieve unprecedented accuracy.  The main shortcoming of DVO is
that it is restricted to the use of depth/intensity image pairs.  In
its original formulation DVO does not naturally incorporate additional
structural cues such as normals or curvature.  The use of these cues
has been shown to substantially enlarge the basin of convergence and
reduce the number of iterations needed to find a solution.

The main contribution of this paper is a general methodology to
photometric sensor data registration that works on cues such as color,
depth, and normal information in a unified way and does not require an
explicit data association between features, 3D points, or
surfaces. The approach is inspired by DVO~\cite{kerl2013icra}.  It
does not require any feature extraction, and operates directly on the
image or image-like data obtained from a sensor such as the Kinect or
a 3D \lidar.  A key property of our approach is an easy-to-extend,
mathematically sound framework for registration that does not need to
make an explicit data association between sensor readings or the 3D
model. In contrast, it solves the registration problem as a
minimization in the color, depth, and normal data exploiting
projections of the sensor data. It can be seen as a generalization of
DVO to handle arbitrary cues and multiple sensing modalities in a
flexible way. An example of this registration is illustrated in
\figref{fig:motivation}. We provide an open source C++ implementation
at:
\begin{quote}
{\footnotesize \url{https://gitlab.com/srrg-software/srrg_mpr}}
\end{quote}

In sum, we make the following key claims:
Our approach
(i) is a general methodology for photometric registration that works transparently with different sensor cues
and avoids an explicit point-to-point or point-to-surface data association,
(ii)  can accurately register typical sensor cues such as \RGBD Kinect or \lidar data exploiting the
 color, depth, and normal information,
(iii) robustly computes the transformation between view points  under realistic
disturbances of the initial guess, and
(iv) can be executed fast enough to allow for online processing of the sensor data without sensor-specific optimizations.
These four claims are backed up through the experimental evaluation.

\section{Related Work}
\label{sec:related}

There exist a large number of different registration approaches.
One general way for aligning 3D point clouds is the ICP algorithm,
which has often been used in the context of range data.
Popular approaches use ICP together with
point-to-point or point-to-plane correspondences~\cite{jaimez2015tro} and
generalized variants such as GICP~\cite{segal2009rss}. There exist approaches exploiting
normal information such as NICP~\cite{serafin2017ras} as well as global
approaches~\cite{yang2016pami} that use branch-and-bound technique
coupled with standard ICP formulation.  A popular and effective approach is
LOAM~\cite{zhang2013rss, zhang2016auro} by Zhang and Singh that
extracts distinct surface and corner features from the point cloud and
determine plane-to-point and line-to-point distances to a voxel-grid
representation.


Traditional approaches to visual odometry track sparse features in
monocular images or stereo pair to estimate the relative orientation
of the images~\cite{forster2017tro, nister2004cvpr}. To deal with
outliers in the data association between feature points, most
approaches use RANSAC to identify inlier and outlier points, combined
with a tracking over multiple frames.
Other approaches rely on a prior for the motion estimate, such as
constant motion model. In presence of external
sensors, such as an IMU, the measurements can be filtered with the achieved
motion estimate~\cite{bloesh2015iros, zheng2017icra}.

Another group of approaches exploits the depth data from \RGBD streams
to register scans and build dense models of the scene. KinectFusion by
Newcombe et al.~\cite{newcombe2011ismar}, for example, largely
impacted the \RGBD SLAM developments over the last 6 years. Similar to
Newcombe et al., the approach of Keller et al.~\cite{keller2013threedv}
uses projective data association for \RGBD SLAM in a dense model and
relies on a surfel-based map~\cite{weise2011cviu} for
tracking. Similar approaches exploit the \RGBD streams by defining
signed distance fields where a direct voxel-based difference is
computed to perform the motion estimation~\cite{slavcheva2016eccv},
making intense use of both CPU and GPU parallelization.
ICP is a frequently used approach
for \RGBD data and special variants for denser depth images have been
proposed~\cite{serafin2015iros, serafin2017ras}.
Recently, the team around Daniel Cremers has proposed semi-dense
approaches using image data~\cite{engel2014eccv} to solve the visual
odometry and SLAM problem as well as dense approaches for featureless
visual odometry for \RGBD data~\cite{kerl2015iccv,kerl2013iros,kerl2013icra}.

In this paper we propose a general and easy-to-implement methodology
for multi-cue photometric registration of 3D point clouds that can be
seen as an extension of DVO.
In contrast to nearly  all previous works, our method has been designed without
considering a specific sensor, nor a particular cue, as we aim to apply the
same exact algorithm in several contexts.

\section{Approach}
\label{sec:approach}

Our approach seeks to register either two observations with respect to each other or an observation against a 3D model.
The sensor observations are assumed to have a 2D representation~$\I$ such as an image
from a regular camera, a range image from a depth camera or a 3D \lidar, or a similar type of observation.
Such a 2D measurement can be seen as an image, where each pixel in the image plane contains one or
more channels, i.e.,  $\I=\{\IChannel \}$
with the channel index~$c$. Examples of such channels are light intensity, depth information, or surface normals.

We aim at registering the current observation to a model, for which 3D information
is available in form of a  point cloud. This model can be a given 3D model, or a point cloud estimated from
the previous observation(s).  We refer to it as the model cloud $\cM=\{ \bp \}$.
In addition to the 3D coordinates, each point
$\bp \in \cM$ can also store multiple cues such as light intensity or a surface normal.

 \begin{figure*}[t]
  \centering
  \includegraphics[width=.85\linewidth]{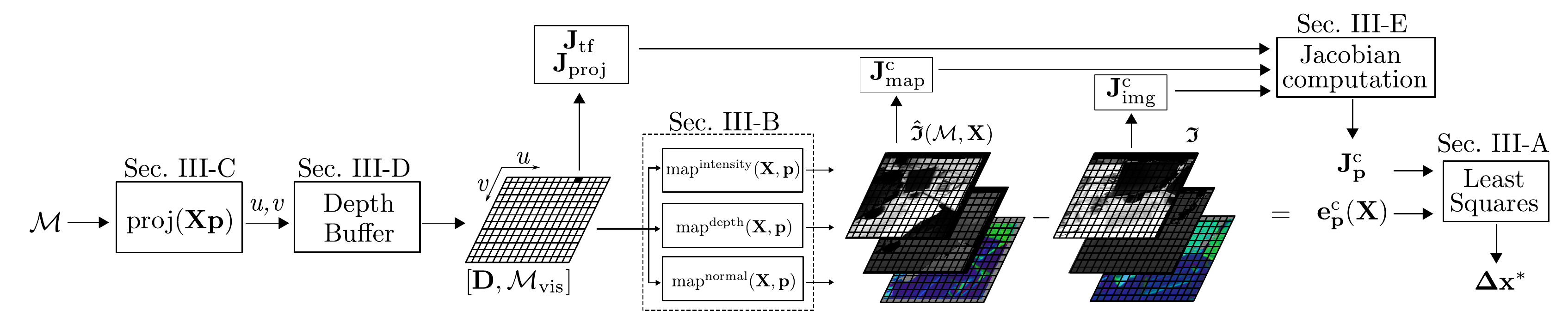}
  \caption{Key ingredients of our framework and links to the corresponding subsections.}
  \label{fig:bigpicture}
\end{figure*}

The following subsections describe the key ingredients of our approach, see also \figref{fig:bigpicture} for an illustration.
\secref{sec:errormin} presents the overall error minimization formulation that uses three functions, which are sensor and/or cue-specific and
\emph{must be implemented by the user} when adding a new cue or a different
sensor, everything else is handled by our framework. These functions are:
\begin{itemize}
	\item \secref{sec:cue-mapping-function}: Cue-specific mapping function $\MapFuncName()$  that describes if and how a cue is transformed through a coordinate transformation.
	\item \secref{sec:projective-models}: Projection model $\ProjFuncName()$ of the sensor, e.g., the pinhole camera model
	\item \secref{sec:jacobians}: The corresponding Jacobians
\end{itemize}

\subsection{Photometric Error Minimization}
\label{sec:errormin}
As in photometric error minimization approaches, our method seeks
to iteratively minimize the pixel-wise difference between the current image~$\I$
and the predicted image~$\IPred$. Here, $\IPred$ is a multi-channel image
obtained by projecting the model $\cM$ onto a virtual camera located at
a pose~$\bP_\bX$, where~$\bX$ is a transformation matrix that transforms the
points of $\cM$ from the global into the local camera coordinate system.
More formally, our method seeks to minimize
\begin{eqnarray}
  \label{eq:total-error}
  \bX^*  &=& \argmin_\bX \sum_{u,v,\channel}
    \|
      \underbrace{
      \IChannelPred[u,v] - \IChannel[u,v]
      }_{\ErrorPixel}
    \|^2_{\bOmega^\channel} \\
  &=& \argmin_\bX \sum_{u,v,\channel}
      {\ErrorPixel}^{\Trans} \, \bOmega^\channel \, \ErrorPixel 
\end{eqnarray}
where $\ErrorPixel$ denotes an error at a pixel location~$\UV$ between the predicted
value~$\IChannelPred[u,v]$ and the measured one~$\IChannel[u,v]$ for a particular
channel index~$c$. The matrix~$\bOmega = \mathrm{diag}(\{{\bOmega^\channel}\})$ is a
block diagonal information matrix used to weight the different channels of the image.

Let~$\ModelVis$ be the subset of points from the model~$\cM$ that are visible from
the image plane of  image~$\I$. In our current implementation, we construct this set
using the depth buffer as explained in details
in~\secref{sec:depth-range-buffer}. Given $\ModelVis$, we can
rewrite the sum in~\eqref{eq:total-error} using the points:
\begin{eqnarray}
  \label{eq:total-error-points}
  \bX^*
  &=& \argmin_\bX \sum_{\channel,\bp \in \ModelVis} \left \| \ErrorPoint^{\Trans}  \right \|^2_{\bOmega^\channel}.
\end{eqnarray}
Each point-wise error term $\ErrorPoint^{\Trans}$ is the difference between a predicted and a
measured channel evaluated at the pixel where the point~$\bp$ projects onto given the model of the sensor.
We expand the point-wise error as follows:
\begin{eqnarray}
  \ErrorPoint
    &=& \MapFunc - \IChannel[\ProjFunc].
    \label{eq:error-term}
\end{eqnarray}
The term $\ProjFunc = \UV$ is a function that computes the image
coordinates obtained by projecting the point~$\bp$ onto a camera
located at~$\bP_\bX$. The function~$\MapFunc$ computes the
\emph{value} of channel $\IChannelPred$ evaluated at the pixel
$\ProjFunc$. During a first read, that may sound confusing as the
default cue light intensity is not affected by
the transformation and are simply copied from the information
in the point $\bp$.  Other cues, however, such as normals or depth values are viewpoint-dependent
and therefore change depending on the given camera pose~$\bP_\bX$.

Our approach minimizes~\eqref{eq:total-error-points} by using a
regularized least squares optimization
procedure. Combining~\eqref{eq:total-error-points}
with~\eqref{eq:error-term} and adding a per-point regularization
weight~$w_\bp$, we can rewrite~\eqref{eq:total-error} in terms of
points as

\begin{small}
  \begin{equation}
    \label{eq:objective-function-expanded}
    \bX^* = \argmin_\bX \sum_{\bp \in \ModelVis} w_\bp \sum_{\channel}  \| \MapFunc - \IChannel[\ProjFunc] \|^2_{\bOmega^\channel},
  \end{equation}
\end{small}
\noindent
where the regularization weight~$w_\bp$  decreases with the
magnitude of the channel errors $\ErrorPoint$ and is used to reject outliers.
The minimization is performed using a local perturbation:
\begin{equation}
\label{eq:perturbation}
\bDeltax = (  \underbrace{\Delta t_x,  \Delta t_y , \Delta t_z}_{\bDeltat} , \underbrace{\Delta
    \alpha_x , \Delta \alpha_y , \Delta \alpha_z}_{\bDelta\boldsymbol\alpha}  )^\Trans,
\end{equation}
consisting of a translation vector $\bDeltat$ and three Euler angles
${\bDelta\boldsymbol\alpha}$. The vector~$\bDeltax$ is a minimal representation
for the transformation matrix~$\bX$ and~\eqref{eq:total-error-points} can be
reduced to a quadratic problem in $\bDeltax$ by computing the Taylor expansion
of~\eqref{eq:error-term}
around a null perturbation as follows:

\begin{small}
\begin{eqnarray}
  \ErrorPoint[\bX \oplus \bDeltax]
  &\simeq& \ErrorPoint + \left. \frac{\partial \ErrorPoint[\bX \oplus \bx]}{\partial \bx}\right|_{\bx = \bZero} \bDeltax  \label{eq:jacobian}\\
  &=&\underbrace{\ErrorPoint}_{\ErrorPointConstant} + \underbrace{\Jp}_{\JpConstant} \bDeltax \label{eq:channel-error}
  \hspace{1.5mm}=\hspace{1.5mm}\ErrorPointConstant + \JpConstant \bDeltax
\end{eqnarray}
\end{small}
The $\oplus$ operator is the transform composition defined as
\begin{eqnarray}
  \label{eq:boxplus}
  \bX \oplus \bDeltax &=&
  \underbrace{
  \begin{bmatrix}
    \bDeltaR&\bDeltat\\
    \bZero & 1
  \end{bmatrix}
  }_{\bDeltax}
  \underbrace{
  \begin{bmatrix}
    \bR&\bt\\
    \bZero & 1
  \end{bmatrix}
  }_{\bX},
\end{eqnarray}
with $\bDeltaR$ obtained by chaining rotations along the three axes:
\begin{equation}
\bDeltaR = \bR_x(\Delta\alpha_x) \bR_y(\Delta\alpha_y) \bR_z(\Delta\alpha_z).
\end{equation}

Thus, the quadratic approximation of~\eqref{eq:objective-function-expanded} becomes
\begin{equation}
  \label{eq:objective-function-linearized}
  \bDeltax^* = \argmin_\bDeltax \sum_{\bp \in \ModelVis} w_\bp \sum_{\channel \in \Channels }  \| \ErrorPointConstant + \JpConstant \bDeltax \|^2_{\bOmega^\channel},
\end{equation}
and $\bDeltax^*$ can be found by solving a linear system
of the form $\bH \bDeltax^* = \bb$ with the term $\bH$ and $\bb$ given by
\begin{eqnarray}
  \bH &=& \sum_{\bp \in \ModelVis} w_\bp \sum_{\channel \in \Channels } {\JpConstantTrans} \bOmega^\channel \JpConstant \\
  \bb &=& \sum_{\bp \in \ModelVis} w_\bp \sum_{\channel \in \Channels } {\JpConstantTrans} \bOmega^\channel \ErrorPointConstant.
\end{eqnarray}

To limit the magnitude of the perturbation between iterations and thus
enforcing a smoother convergence, we solve a damped linear system of the form
\begin{equation}
\label{eq:linsys}
\left (\bH + \lambda \bI \right) \bDeltax^* = \bb.
\end{equation}

Solving~\eqref{eq:linsys} yields a perturbation $\bDeltax^*$ that
minimizes the quadratic problem. A new solution $\bX^*$ is found by
applying $\bDeltax^*$ to the previous solution $\bX$ as
\begin{equation}
\bX^* = \bX \oplus \bDeltax^*.
\end{equation}

This subsection provided the overall minimization approach and in the remainder
of this section, we describe the parts of our
approach in detail. These details are the cue-specific mapping function and the projection model.
Furthermore, we explain how to  use the depth buffer to construct $\ModelVis$ and finally
discuss the structure of the Jacobians and a pyramidal approach to the optimization.

\subsection{Cue Mapping Function}
\label{sec:cue-mapping-function}

In contrast to traditional ICP approaches but similar to
DVO~\cite{kerl2013icra}, our method does not use an explicit point-to-point
data association procedure. In addition to that, by abstracting the
cues into channels of the image, our method can be extended to deal
with an arbitrary number of cues and can benefit from all the
available information. In our current implementation, we consider the
following cues: \textit{intensity}, \textit{depth}, \textit{range} and
\textit{normals}. Note that additional cues or further sensor information
can be added easily without changing the framework.

In this subsection, we present the mapping functions~$\MapFuncName()$ for the
used cues {intensity}, {depth}, {range}, and {normals} that depend on~$\bX$. We
will use $\bR$ and $\bt$ as defined in~\eqref{eq:boxplus} as the rotation matrix
and translation vector of~$\bX$.

\textbf{Intensity:} The intensity is not a \emph{geometric} property of the point
and thus it is not affected by the transformation defined through $\bX$, therefore the intensity
value of a point $\intensity(\bp)$ is considered invariant under the
map function, i.e.,
\begin{equation}
  \MapFunc[\intensity] = \intensity(\bp).
\end{equation}

\textbf{Depth:} The depth cue of a point is the $z$-component of the
point transformed by $\bX$, i.e.,
\begin{equation}
  \MapFunc[\depth] = \begin{bmatrix}0 & 0 & 1\end{bmatrix} (\bR \bp + \bt).
\end{equation}

\textbf{Range:} The range cue of a point is the norm of the point
transformed by $\bX$,  i.e.,
\begin{equation}
  \MapFunc[\range] = \| \bR \bp + \bt \|.
\end{equation}

\textbf{Normals:} The normal cue of a point~$\bp$ is the normal vector specified by $\bn(\bp)$ at the point
\emph{rotated} by $\bX$,  i.e.,
\begin{equation}
  \MapFunc[\normal] = \bR \, \bn(\bp).
\end{equation}

\subsection{Projection Models}
\label{sec:projective-models}

The projection function maps a 3D point from the
model cloud onto a coordinate in a 2D image. In our implementation, we provide two
projective models, the \emph{pinhole} and the \emph{spherical} model. The pinhole model better
captures the characteristics of imaging sensors such as \RGBD cameras while the
spherical model is entailed to 3D \lidar{}s.
In the paper, we describe these two models but note that the framework easily extends to other types of
projection functions such as the cylindrical model. Only a single function needs to be overridden.

\textbf{Pinhole Model:} Let $\bK$ be the camera matrix. Then, the pinhole
projection of a point $\bp$ is computed as
\begin{eqnarray}
  \ProjFuncChannel[\pinhole] &=& \pi(\bK \, \bp)\\
  \bK&=&\begin{bmatrix}
    f_x& 0 & c_x\\
    0 & f_y & c_y\\
    0 & 0 &1
  \end{bmatrix} \label{eq:camera-matrix}\\
  \pi(\bv) &=& \frac{1}{v_z}
  \begin{bmatrix}
    v_x \\ v_y
  \end{bmatrix}
  \label{eq:pinhole-projection},
\end{eqnarray}
with the intrinsic camera parameters for the focal length~$f_x$, $f_y$ and the principle point~$c_x$, $c_y$. The function~$\pi(\bv)$ is the homogeneous normalization.

\textbf {Spherical Model:} Let $\bK$ be a camera matrix in the form of~\eqref{eq:camera-matrix}, where $f_x$ and $f_y$
specify respectively the resolution of azimuth and elevation and $c_x$ and $c_y$
their offset in pixels. Then, the spherical projection of a point is given by
\begin{eqnarray}
  \ProjFuncChannel[\spherical] &=& \bK
  \begin{bmatrix}
    \atantwo(\bp_y, \bp_x) \\
    \atantwo\left(\bp_z, \sqrt{\bp_x^2 + \bp_y^2}\right)\\
    1
  \end{bmatrix}\label{eq:spherical-projection}
\end{eqnarray}

\subsection{Computing Visible Points using Depth and Range Buffers}
\label{sec:depth-range-buffer}
As stated in the beginning of~\secref{sec:approach}, \eqref{eq:total-error}
and \eqref{eq:total-error-points} are equivalent if there are no occlusions or self-occlusions.
This condition can be satisfied by removing all points from the model $\cM$ that
would be occluded after applying the projection. At each
iteration, the estimated position $\bX$ of the cloud with respect to the sensor
changes, thus before computing the projection, we need to transform the model
according to $\bX$. Subsequently, we project each transformed point onto an
image and for each pixel in the image, we preserve only the point that is
the closest one to the observer according to the chosen projective model. The outcome of
the overall procedure is a subset of the transformed points in the model that
are visible from the origin.

The reduced set of non-occluded points can be effectively computed as follows.
Let $\bD$ be a 2D array of the size of the image, each cell $\UV$
of $\bD$ contains a depth or range value referred to
as $\bD(u,v)$ and a  model point $\hat\bp$ that generated this depth or range reading.
 First, all depths $\bD(u,v)$ are  initialized with $\infty$.
Then, we iterate over all points $\bp \in \cM$ and perform the following computations:
  \begin{itemize}
  \item Let $\bp'=\bR \bp+\bt$ be the transformed point and let
    $(u,v)=\ProjFunc[\bp']$ be the image coordinates of the point after
    the projection.
  \item If using the pinhole projective model, let $d'=\bp'_z$ be the $z$
    component of the transformed point. If using the spherical
     model, let $d'=\|\bp'\|$ be its norm.
  \item We compare the range value $\bD(u,v)$ previously stored in $\bD$
    with the range computed for the current point $d'$. If the latter
    is smaller than the former, we replace $\bD(u,v)$ with $d'$ and
    $\hat\bp$ with $\bp'$.
  \end{itemize}

At the end of the procedure, $\bD$  contains all points of the model
visible from the origin, i.e.~are not occluded. These points form the $\ModelVis$ cloud.

\subsection{Structure of the Jacobian}
\label{sec:jacobians}

In this section, we highlight a modular structure of the Jacobian $\Jp$, which
is key for an efficient computation. By applying the chain rule to the right
summand of~\eqref{eq:jacobian}, we obtain the following form for the Jacobian $\Jp$:

\begin{scriptsize}
  \begin{eqnarray}
    \Jp &=& \overbrace{\frac{\partial \MapFuncDer}
      {\partial \bx}\bigg|_{\bx = 0}}^{\Jmap}  \\
    && -\underbrace{\frac{\partial \IChannel[u,v]}
      {\partial u,v}\bigg|_{u,v = \ProjFunc[\bX\bp]}}_{\Jimg}
    \underbrace{\frac{\partial \proj(\brevep)}{\partial \brevep}\bigg|_{\brevep = \bX\bp}}_{\Jproj}
    \underbrace{\frac{\partial (\bX \oplus \bx) \, \bp }{\partial \bx}\bigg|_{\bx = 0}}_{\Jtf} \nonumber
  \end{eqnarray}
\end{scriptsize}
Thus, the Jacobian can be compactly written as:
\begin{eqnarray}
  \label{eq:jacobian_structure}
  \Jp = \Jmap - \Jimg \, \Jproj \, \Jtf
\end{eqnarray}
Note that the multiplicative nature of
\eqref{eq:jacobian_structure} in this formulation allows us to easily compute the
overall Jacobian from its individual components $\Jmap$, $\Jimg$, $\Jproj$ and
$\Jtf$.
In particular, $\Jtf$ does not depend on the channel, nor on the projective
model. Similarly, $\Jproj$ depends only on the projective model.  The image
Jacobian $\Jimg$ can be computed directly from the image through a convolution
for obtaining image gradients. Note that
to increase the precision, we recommend to compute $\Jimg$ with subpixel precision
through bilinear interpolation during the optimization.  Only the Jacobian $\Jimg$ of the function
$\map(\cdot)$ depends on the specific cue.  For completeness, we report all
the Jacobians for all projective models and all channels in Appendix~\ref{appendix:jacobian}.

\subsection{Hierarchical Approach}



A central challenge for photometric minimization approaches is the choice
of the resolution in order to optimize the trade-off between the size
of the convergence basin and the accuracy of the solution.
A high image resolution has a positive effect on the accuracy
of the solution given the initial guess lies in the convergence basin.
This is due to the fact that more measurements are
taken into account and the precision of the typical sensor is
exploited in a better way.
A high resolution, however, often comes at the cost of reducing the convergence
basin since the iterative minimization can get stuck more easily
in a local minimum arising from a high frequency spectral
component of the image. Thus, using lower resolution, the photometric
approach exhibits an increased convergence basin at the cost of a
lower precision.

In our implementation, we leverage on these
considerations by using a pyramidal approach. The optimization is
performed at different resolutions, starting from  low to
high resolutions. After convergence on one level, the optimization switches to the next
level by increasing the resolution. The optimization on the higher resolution
uses the solution from the lower level as the initial guess.
Our termination criterion for the optimization on each level of this resolution pyramid
analyzes the evolution of the value of the objective function
in~\eqref{eq:total-error-points}, normalized by the number of inliers.
We stop the iterations at one level if this value does not
decrease between two subsequent iterations or if a maximum number of
iterations is reached.

\subsection{Brief Summary}
As can be seen from the overall \secref{sec:approach}, we provide a general methodology
that builds upon photometric registration and works flexibly with different sensor cues.
All that is needed for integrating a new sensor cue is  an implementation of the mapping function~$\MapFuncName()$,
the projection function~$\ProjFuncName()$, and the Jacobians (see Appendix).
Furthermore, our method does not require any
explicit point-to-point or point-to-surface correspondence as the optimization
performs the error minimization exploiting the projections. Thus, the first of our
four claims made in the introduction is backed up through  \secref{sec:approach}.

\section{Experimental Evaluation}
\label{sec:exp}

Our method is a general and efficient framework for multi-cue sensor data registration.
We release a C++ implementation that closely follows the description in this paper as open source.
We implemented our general methodology for the \RGBD Kinect sensor and 3D laser scanners
using the following cues: intensity, depth, range, and surface normal and thus
the evaluation is done based on these sensors and cues.
 Note,  that further cues or similar sensors can be added easily.

The evaluation presented here is designed to support the remaining three claims made in the introduction.
To simplify comparisons, we conducted our experiments on publicly available datasets:
\begin{itemize}
\item TUM benchmark suite~\cite{sturm2012iros}, acquired with \RGBD sensors in office-like environments.
\item S.~Gennaro Catacomb dataset~\cite{rovina2016aich}, recorded with a RobotEye 3D \lidar in a
  catacomb environment.
\item KITTI dataset~\cite{geiger2012cvpr}, where we used the Velodyne HDL-64E data recorded in large scale environments.
\end{itemize}
Furthermore, we provide  comparisons to state-of-the art approaches such as DVO or NICP for each dataset.
The outcomes of these comparisons highlight that our method, although being general and relatively easy to
implement,  yields to an accuracy that is comparable to those achieved by systems dedicated
 to specific setups.

\subsection{Registration Performance and Comparison}

\begin{table*}
  \caption{Relative Pose Error on TUM {desk} sequences.}
  \centering
      {\footnotesize
        \begin{tabular}{|l|c|c|c|c|c|c|c|c|}
          \hline
          Approach / setup & \multicolumn{2}{c|}{fr1/desk2} & \multicolumn{2}{c|}{fr1/desk} & \multicolumn{2}{c|}{fr2/desk} & \multicolumn{2}{c|}{fr2/person}\\
          & [m/s]  & [deg/s]& [m/s]  & [deg/s] & [m/s]  & [deg/s] & [m/s]  & [deg/s]\\ \hline
          DVO (as reported in paper)    & 0.0687 &  -  & 0.0491 &  -   & 0.0188 &  -  & 0.0345 &  -  \\ \hline
          DVO (implementation)           & 0.0700 & 5.14   & 0.0580 & 3.83    & 0.0318 & 1.15    & 0.0360 & 0.99    \\ \hline
          Our approach   & 0.0920 & 5.14   & 0.0614 & 3.32    & 0.0365 & 1.65    & 0.0481 & 1.45     \\ \hline
\hline
          DVO (without intensity cue) & - & -  & - & -  & - & -  & - & -   \\ \hline
          Our approach (without intensity cue) & 0.1073 & 5.20  & 0.0630 & 3.66  & 0.0329 & 1.52  & 0.0411 & 1.27   \\ \hline
        \end{tabular}
      }
      \label{tab:tum_results}
\end{table*}

This section is designed to show that our method can accurately register typical sensor cues such as \RGBD Kinect or \lidar data exploiting the  color, depth, and normal information and that it can do so under realistic disturbances of the initial guess.
The first experiment is designed to evaluate the accuracy of our
approach.  To do that, we rely on  \RGBD{} data and provide a comparison to
 Dense Visual
Odometry (DVO)~\cite{kerl2013icra}, which is the current state-of-the-art method
and the one most closely related to this paper.
We used  the three available
channels, namely the {intensity}, the {depth} and the
{normals}.

We performed the comparison on the four {desk} sequences also used
in~\cite{kerl2013icra} and report the relative pose error (RPE). For
DVO, we used the author's open source
implementation\footnote{\url{https://github.com/tum-vision/dvo_slam}}.
For completeness, we report in
\tabref{tab:tum_results} both, the results presented in the paper~\cite{kerl2013icra}
and the ones we obtained with the open source
implementation.  All values have been computed using the evaluation
script provided with the TUM benchmark suite.  Our approach provides
slightly lower but overall comparable
performance than DVO yielding a  low relative error
both, in terms of translation and rotation,
respectively in the order of $10^{-2}$\,m/s and 1\,deg/s, see \tabref{tab:tum_results}.

We conducted a second experiment with the TUM dataset, where we
removed the intensity channel, using only the depth images
and the normals derived from the depth image to perform the registration.
By exploiting
the normal cue, our approach provides results consistent to the ones
obtained when using also the intensity, see last two rows on
\tabref{tab:tum_results}.  In contrast to that, DVO was unable to perform the
registration without the intensity channel.  This is coherent with the
operating conditions which DVO was designed for and at the same
time supports the general applicability of our method to different cues.

The third experiment aims at showing the effectiveness of our
algorithm when dealing with dense 3D laser data.  We used a dataset
acquired in the S.~Gennaro catacomb of Naples within the EU project
ROVINA.  The data was recorded with a Ocular RobotEye RE05 3D laser scanner using a
maximum range of 30\,m.  Since the dataset does not provide
ground truth, we performed a qualitative comparison with Normal ICP
(NICP) by Serafin et al.~\cite{serafin2017ras}.

The 3D point clouds have been recorded in a stop and go fashion at
an average distance of about 1.7\,m between two consecutive scans. The robot has tracks and
this provides a rather poor odometry information. This odometry is used as the initial guess for the
optimization and the same guess was also used for NICP, that is used for comparisons in this experiment.
To perform the registration we use both, {range} channel and the
{normals} channel computed on the range image. There is no intensity information available for this dataset.

\begin{figure}
  \centering
  \includegraphics[width=1\linewidth]{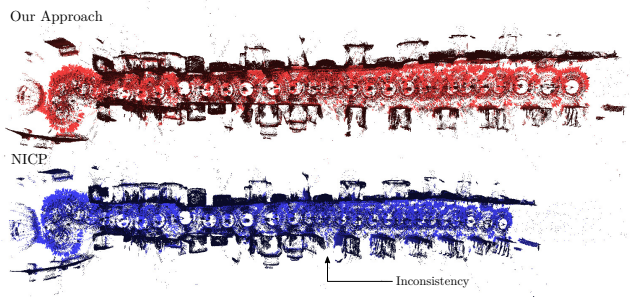}
  \caption{{S.~Gennaro} catacombs dataset, recorded with a
    RobotEye 3D \lidar. Direct comparison of our approach with Normal
    ICP (NICP). The latter shows a registration inconsistency in the middle of the trajectory, with an unexpected change of the orientation around the vertical axis.}
  \label{fig:san_gennaro_comparison}
\end{figure}

\begin{figure}
  \centering
  \includegraphics[width=1\linewidth]{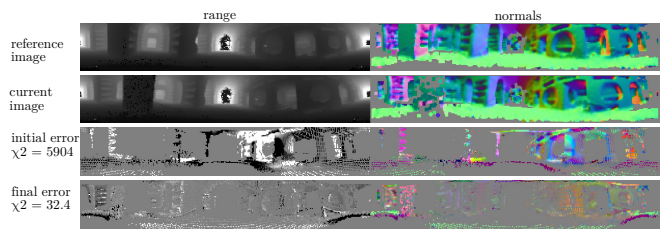}
  \caption{Illustration of the error reduction while registering two images of the
    {S.~Gennaro} dataset (best viewed on screen).}
  \label{fig:san_gennaro_error_images}
\end{figure}

\begin{figure}
  \centering
  \includegraphics[width=1\linewidth]{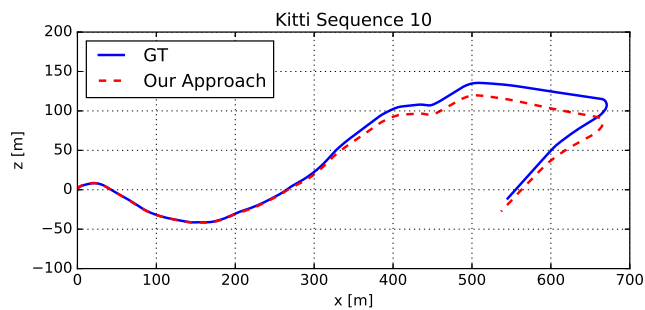}
  \caption{Ground truth comparison in the sequence 10 of the KITTI
    dataset.}
  \label{fig:kitti_seq_10}
\end{figure}

\begin{figure}
  \centering
  \includegraphics[width=1\linewidth]{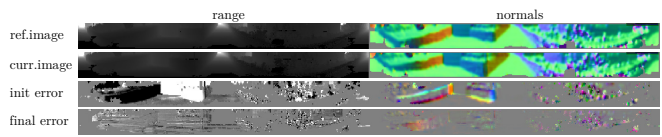}
  \caption{Illustration of the error before and after registering two
    images of the KITTI dataset. The images are obtained by projecting
    the Velodyne HDL-64 clouds using a spherical projector  (best viewed on screen).}
  \label{fig:kitti_seq_10_error_images}
\end{figure}

Fig.~\ref{fig:san_gennaro_comparison} illustrates the two
reconstructions of one of the sequences of the {S.~Gennaro} catacombs,
while Fig.~\ref{fig:san_gennaro_error_images} shows the error of a
pairwise registration before and after the alignment.  Thanks to the
abstraction provided by the map function and the projective model (see
\secref{sec:cue-mapping-function} and~\ref{sec:projective-models}),
our method can deal with both \RGBD data and 3D scans in a uniform manner.

To further stress the generality of our method, we conducted an
additional experiment using exactly the same code and parameters using
the sequence 10 of the KITTI dataset~\cite{geiger2012cvpr} where we
used the 3D scans obtained with a Velodyne HDL-64E \lidar.  As in the
previous experiment, the {range} and {normals} cues are used.
\figref{fig:kitti_seq_10_error_images} illustrates the error reduction
after the registration of two scans.

As shown in \figref{fig:kitti_seq_10}, our output reflects the ground
truth with an error of the 7\% of the trajectory length, with a
rotational error of 0.025\,deg/m.  Albeit reasonable, this accuracy
is still below the one provided by approaches dedicated to the sparse
\lidar such as LOAM~\cite{zhang2016auro}.  We see the reason of this
lower performance in the quantization effects affecting the
projections when the clouds are sparse. We plan to address this aspect
in future versions by using anisotropic projection functions.
Moreover, approaches such as LOAM take advantage from edge and planar
surface features, whose usage particularly helps in cases of structure
lack.

\subsection{Runtime}

The next set of experiments is designed to support the last claim,
namely that our approach can be executed fast enough to allow for
online processing of the sensor data without sensor-specific
optimizations.  Thus, we report in the remainder of this section
the runtime statistics.  We performed all the presented experiments on
two different computers running Ubuntu 16.04. One is a laptop
equipped with a i7-3630MQ CPU with 2.40\,GHz and the second one is a
desktop computer equipped with a i7-7700K CPU with 4.20\,GHz.
Our software runs on a single core and in a single thread.

\tabref{tab:speed} summarizes the runtime results for the different
configurations presented above. We provide the average
results obtained in the four TUM {desk} sequences for the Kinect \RGBD
and depth-only configurations, as well as for the two \lidar setups.
As listed in the table, our method can be executed fast and in an
online fashion.  On a mobile i7 CPU, we achieve average frame rates of
20\,Hz with Kinect \RGBD sensor and 14\,Hz with the Velodyne HDL-64E
data, while we achieve average of 30\,Hz and 23\,Hz on an i7 desktop
computer for the same configurations.
Furthermore note, that our approach has a small memory footprint. For
the whole registration procedure of the three datasets, we always required
less than 200\,MB (Kinect: 178\,MB; RobotEye: 140\,MB; HDL-64E: 198\,MB).

Note that recording a single RobotEye point clouds takes around 30\,s,
i.e.  the robot stops, records, and then restarts.  Thus, this setup
may not be considered for real-time usage. In contrast, the
Velodyne clouds of the KITTI dataset have been recorded at an average
frame rate of 10\,Hz and the data can be processed online.

\begin{table}
  \caption{Average image processing runtime with std. deviation}
  \centering
      {\footnotesize
        \begin{tabular}{|c|c|c|}
          \hline
          Sensor & laptop computer & desktop computer\\
          & i7-3630MQ 2.4\,GHz & i7-7700K 4.2\,GHz  \\
          \hline
          Kinect     & 49\,ms~$\pm$~0.8\,ms~$\approx$~20.1\,Hz
          & 33\,ms~$\pm$~0.6\,ms~$\approx$~29.8\,Hz
           \\
          RobotEye    & 324\,ms~$\pm$~1.5\,ms~$\approx$~3.1\,Hz
          & 250\,ms~$\pm$~0.8\,ms~$\approx$~4\,Hz
            \\
          HDL-64E  & 67\,ms~$\pm$~0.3\,ms~$\approx$~14.8\,Hz
          & 43\,ms~$\pm$~0.1\,ms~$\approx$~23.1\,Hz
            \\
          \hline
        \end{tabular}
      }
      \label{tab:speed}
\end{table}

In summary, our evaluation suggests that our method provides competitive results  in
several different scenarios compared to  approaches dedicated to a specific setup.
At the same time, our method is fast enough for online processing and has small memory
demands, proportional to the number of channels in use.  Thus, we conclude that we supported
all our four claims made in the introduction.

\section{Conclusion}
\label{sec:conclusion}

In this paper, we presented a general framework for registering sensor
data such as \RGBD or 3D \lidar data.  Our approach extends dense
visual odometry and operates on the different available cues such as
color image, depth, and normal information. Our method avoids an
explicit data association and operates by direct error minimization
using projections of the sensor data or model.  This allows us to
successfully register data effectively without tricky, sensor-specific
adaptations. We implemented and evaluated our approach on different
datasets and provided comparisons to other existing techniques and
supported all claims made in this paper. The experiments suggest that
we can accurately register \RGBD and 3D data under realistic
configurations and that the computations can be executed at the sensor
framerate on a regular notebook computer using a single core.

\appendices

\section{Jacobians}
\label{appendix:jacobian}
In this section, we focus in more detail on each part of the Jacobian
in~\eqref{eq:jacobian_structure} and specify all terms used of this equation.

\subsection{Jacobian of Transformation}
\label{appendix:jacobian_icp}
We are using the $\vectorToTransform(\cdot)$
operator to denote the conversion between $\bX$ and $\bDeltax$ as defined
in \eqref{eq:perturbation}. By applying the $\vectorToTransform(\cdot)$
operator, we obtain the standard Jacobian:
\begin{eqnarray}
  \Jtf = \frac{\partial \vectorToTransform(\bx)\,\brevep}{\partial \bx}\bigg|_{\bx=0} =
  \begin{bmatrix}\bI & -\brevep_{\times} \label{eq:jtf}
    \end{bmatrix}
\end{eqnarray}
where $\bI$ is a $3 \times 3$ identity matrix and $\brevep_{\times}$ denotes the
skew-symmetric matrix formed from $\brevep$.

\subsection{Map Function Jacobian}
The map function Jacobian $\Jmap$ differs for each of the considered channels
$\channel$.  In this work, we use the following channels: \textit{intensity},
\textit{depth}, \textit{range} and \textit{normals}. In the following we present
the Jacobian derivation of each of these cues.

\textbf{Intensity: }
The map function does not affect the intensity, thus we have:
\begin{equation}
\frac{\partial \MapFuncDer[\intensity]}{\partial \bx}\bigg|_{\bx = 0} = 0
\end{equation}

\textbf{Depth: } We have already computed the derivative of the transformation
function $\Jtf$ in \eqref{eq:jtf}. For \RGBD sensors, the depth
is computed as the $z$ coordinate of the transformed point $\bp$. Thus, the map
Jacobian for the depth channel is the third row of $\Jtf$
\begin{equation}
  \frac{\partial \MapFuncDer[\depth]}{\partial \bx}\bigg|_{\bx = 0} =
  \begin{bmatrix}
    0 & 0 & 1
  \end{bmatrix}
  \cdot \Jtf
\end{equation}

\textbf{Range: } When using a 3D \lidar, the range $r = \|\bp\|$ replaces the depth.
Thus, the Jacobian $\bJ_\map^\range$ is computed as
\begin{eqnarray}
  \bJ_\map^\range &=&
  \frac{\partial \MapFuncDer[\range]}{\partial \bx}\bigg|_{\bx = 0} \nonumber\\
  &=&
  \frac{\partial \range(\vectorToTransform(\bx) \, \bX \bp)}{\partial \bx}\bigg|_{\bx = 0} \nonumber\\
  &=&
  \frac{\partial \range(\brevep)}{\partial \brevep}\bigg|_{\brevep =\bX \bp}
  \,
  \frac{\partial \vectorToTransform(\bx)\, \bX \bp}{\partial \bx}\bigg|_{\bx = 0} \nonumber\\
  &=& \frac{\partial \range(\brevep)}{\partial \brevep}\bigg|_{\brevep =\bX \bp} \, \Jtf \nonumber \\
  &=& \frac{1}{\norm(\bp)}
  \begin{bmatrix}
    \bp_x & \bp_y & \bp_z
  \end{bmatrix} \cdot \Jtf
\end{eqnarray}

\textbf{Normals: } The mapping function applied to the normals cue depends on
the rotational part of $\bX$ and the Jacobian is given by
\begin{equation}
  \frac{\partial \map^{\normal}(\bX \oplus \bx, \bp)}{\partial \bx}\bigg|_{\bx = 0} =
  \begin{bmatrix}
    \bZero & -{[\bR \, \bn(\bp)]}_{\times}
  \end{bmatrix},
\end{equation}
where $\bR$ denotes a rotation matrix, $\bn(\bp)$ the normal defined for the point
$\bp$, and ${[\bR \, \bn(\bp)]}_{\times}$  the skew-symmetric matrix.

\subsection{Image Jacobian}
The Image Jacobian $\Jimg$ is numerically computed for
each channel $\channel$ with pixel-wise derivation:
\begin{eqnarray}
  \frac{\partial \IChannel[u,v]}{\partial u} &=& \frac{1}{2}\left(\IChannel[u+1,v] - \IChannel[u-1,v]\right) \nonumber \\
  \frac{\partial \IChannel[u,v]}{\partial v} &=& \frac{1}{2}\left(\IChannel[u,v+1] - \IChannel[u,v-1]\right)
\end{eqnarray}

\subsection{Jacobian of Projection}
The Jacobian of the projection depends directly on the projection function
$\Jproj = \frac{\partial \proj(\bp)}{\partial \bp}$. In the following, we
provide details for the two models given in \secref{sec:projective-models}.

\textbf{Pinhole Model: } We derive the projection Jacobian for a pinhole camera
from~\eqref{eq:pinhole-projection}:
\begin{eqnarray}
  \frac{\partial \proj(\bp)}{\partial \bp} &=& \frac{\partial \pi(\bp')}{\partial \bp'}\bigg|_{\bp'=\bK\,\bp} \, \frac{\partial \bK\,\bp}{\partial \bp} \\
  &=& \frac{1}{z^2}
  \begin{bmatrix}
    z & 0 & -x  \\
    0 & z & -y
  \end{bmatrix}
  \bigg|_{(x,y,z) = \bK\,\bp}
  \, \bK \nonumber
\end{eqnarray}

\textbf{Spherical Model: } The projection function for the range sensors is
defined in \eqref{eq:spherical-projection}. We make use of the substitution
$\ba_2 = \sqrt{\bp_x^2 + \bp_y^2}$, and define the Jacobian as follows:
\begin{eqnarray}
  \frac{\partial \proj(\bp)}
       {\partial \bp} =
       \begin{bmatrix}
         \frac{1}{\ba_2^2}\,\begin{bmatrix}-\bp_y & \bp_x & 0\end{bmatrix}\\
         \frac{1}{\ba_2^2+\bp_z^2}\,\begin{bmatrix}-\frac{\bp_x\, \bp_z}{\ba_2} & -\frac{\bp_y\, \bp_z}{\ba_2} & \ba_2\end{bmatrix}\\
         \begin{bmatrix}0 & 0 & 0\end{bmatrix}
       \end{bmatrix}
       \, \bK
\end{eqnarray}


\bibliographystyle{plain}
\bibliography{glorified.bib}

\end{document}